\documentclass[twoside]{article}

\usepackage[preprint]{neutralpreprint}
\usepackage[round,authoryear]{natbib}
\usepackage{amsmath,amssymb,mathtools}
\usepackage{booktabs}
\usepackage{multirow}
\usepackage{array}
\usepackage{graphicx}
\usepackage{xcolor}
\usepackage{microtype}
\usepackage{url}
\usepackage{enumitem}
\usepackage{tikz}
\usetikzlibrary{arrows.meta,positioning,fit,plotmarks}

\newcommand{\method}{\textsf{ProtocolMatch}}
\newcommand{\mse}{\operatorname{MSE}}
\newcommand{\risk}{\mathcal{R}}
\newcommand{\protocol}{\mathcal{P}}
\newcommand{\dataset}{\mathcal{D}}
\newcommand{\archset}{\mathcal{A}}
\newcommand{\winner}[1]{\textbf{#1}}
\definecolor{datasetblue}{HTML}{3B6D97}
\definecolor{datasetorange}{HTML}{B76A35}
\definecolor{datasetgreen}{HTML}{40836F}

\begin{document}

\runningtitle{ProtocolMatch: Protocol-Dependent Model Selection}
\runningauthor{Lu Wei, Yufeng Wang, and Haibin Ling}

\twocolumn[
\preprinttitle{ProtocolMatch: Protocol-Dependent Model Selection\\for Scientific Dynamics Forecasting}

\preprintauthor{Lu Wei\textsuperscript{*} \And Yufeng Wang\textsuperscript{*} \And Haibin Ling}
\preprintaddress{Stony Brook University \And Stony Brook University \And Westlake University}
\begin{center}
\small\textsuperscript{*}Equal contribution.
\end{center}
]

\begin{abstract}
Scientific dynamics forecasting is often framed as an architecture choice, although deployment is also determined by observed history, rollout feedback, compute budget, physical objective, and test distribution. We formulate protocol-dependent model selection and introduce \method{}, a compute-matched, validation-selected, and failure-preserving evaluation framework. On driven quantum-spin dynamics, we compare recurrent, patched-attention, causal-attention, and low-rank linear predictors across three independently generated datasets. The causal-attention--recurrence ordering reverses as the training set grows within a fixed two-spin task, while a linear predictor has the lowest mean error in the six-spin local-observable comparison. Restricting observed history worsens every refreshed-history view but improves every closed-loop view in the four-spin study. A latest-state MLP has lower error than persistence on every dataset under state refresh across all five cells, yet its closed-loop rank varies by system and includes finite explosive errors. Physical penalties improve targeted consistency without reliably improving prediction error, and in-distribution intervals lose most coverage after a driving-frequency shift. Thus scientific model selection should return a predictor with its protocol and report accuracy, physical validity, and shifted-distribution reliability separately.
\end{abstract}

\section{Introduction}

Scientific forecasting models compress an observed history into a state representation that can be propagated beyond the data seen during training. Architecture determines one part of that representation. The forecasting protocol determines another: which variables are observed, whether later inputs come from new measurements or previous predictions, how much optimization is allowed, and whether physical constraints compete with data fit. A model rank obtained under one such protocol need not survive another.

Quantum dynamics makes the distinction operational. A driven many-body system evolves causally under a time-dependent Hamiltonian, while measurements expose only a chosen set of observables. In an observed-history forecast, each new input window is refreshed from the reference trajectory. In a closed-loop forecast, the model's own output enters the next window. Both protocols use the same known future controls, but they ask different questions: local predictive accuracy versus autonomous stability. Recurrent models have forecast quantum observables under random driving and random-circuit evolution \citep{mohseni2022deep,mohseni2024deep}, and attention-based models have produced long-horizon predictions for dissipative dynamics \citep{herrera2024short}; neither result supplies a protocol-independent ordering.

We therefore study \emph{protocol-dependent model selection}: the comparison target is an architecture evaluated under an explicitly declared forecasting protocol. Our framework, named \method{}, holds candidate capacity and optimization time approximately fixed, selects checkpoints and learning rates using validation data only, treats independently generated datasets as the replication unit, and preserves numerical and infrastructure failures in every denominator. It evaluates prediction error, physical consistency, and interval coverage separately.

The empirical testbed contains driven two-, four-, and six-spin systems. Two- and four-spin tasks expose complete Pauli-observable vectors, whereas the bounded six-spin extension exposes local observables. Four model families span recurrent state tracking, patched attention, causal token attention, and a low-rank linear map. The experiments ask four linked questions: whether architecture ranks change with data and target space, whether the value of observed history changes with rollout, whether physical validity and predictive accuracy move together, and whether in-distribution calibration transfers under frequency shift.

Our contributions are:
\begin{enumerate}[leftmargin=*,itemsep=2pt,topsep=2pt]
    \item We formulate scientific forecasting comparisons around an explicit protocol \(\protocol\), making observation, rollout, clipping, compute, and distributional conditions part of the model-selection estimand.
    \item We develop a controlled evaluation design with parameter matching, optimization-time matching, validation-only selection, independent dataset replication, and failure-preserving aggregation.
    \item We identify replicated ranking reversals across data regimes and rollout rules, then show that prediction error, physical-constraint satisfaction, and shifted-distribution coverage need not move together.
\end{enumerate}

The result is constructive: the experiments do not yield a universal winner, but they do yield conditional selection rules. The protocol must therefore be declared and varied before an architecture rank can be interpreted.

\section{Related Work}

\paragraph{Learning physical dynamics.}
Neural ODEs, Hamiltonian neural networks, and graph simulators encode different assumptions about continuous-time or interacting systems \citep{chen2018neuralode,greydanus2019hamiltonian,sanchezgonzalez2020learning}. Physics-informed learning augments empirical risk with residual or validity penalties \citep{raissi2019physics,karniadakis2021physics}. For quantum systems, learned representations have been used for state approximation, process prediction, control, and Hamiltonian inference \citep{carleo2017solving,huang2023learning,norambuena2024physics,an2025dual}. In contrast, our contribution is not a new simulator; it is a model-selection methodology for comparing forecasting biases under controlled deployment conditions.

\paragraph{Sequence priors for forecasting.}
Long short-term memory networks encode state through gated recurrence \citep{Hochreiter&Schmidhuber97nc}. Transformers instead mix tokens through attention \citep{vaswani2017attention}, and forecasting variants use decomposition, frequency structure, or temporal patches \citep{wu2021autoformer,zhou2022fedformer,nie2022time}. Simple linear predictors can rival larger forecasters on standard time-series benchmarks \citep{zeng2023transformers}. These findings motivate a comparison that includes both causal attention and a linear control, rather than treating one patched Transformer as the entire alternative class.

\paragraph{Benchmark reliability and uncertainty.}
Model ranks can be sensitive to data and optimization variation \citep{bouthillier2021accounting}. In quantum learning, resource-controlled studies likewise show that stronger model classes need not dominate simpler ones \citep{zhao2025rethink}. Standard conformal prediction provides finite-sample coverage under exchangeability, while distribution shift motivates adaptive variants \citep{angelopoulos2023conformal,gibbs2021adaptive}. We instead keep identity-distribution calibration fixed and measure its loss of reliability under a prespecified shift, rather than interpreting in-distribution coverage as evidence of shift robustness.

\paragraph{Positioning.}
Existing benchmark methodology typically treats data splits, hyperparameter search, and random seeds as sources of uncertainty around a single task. Our setting adds a second layer: changing the observation set or rollout rule changes the conditional distribution seen by the predictor. We make that layer explicit and test whether conclusions survive across protocols. This differs from proposing another forecaster and from declaring a winner after aggregating unlike deployment conditions.

\section{Protocol-Dependent Model Selection}

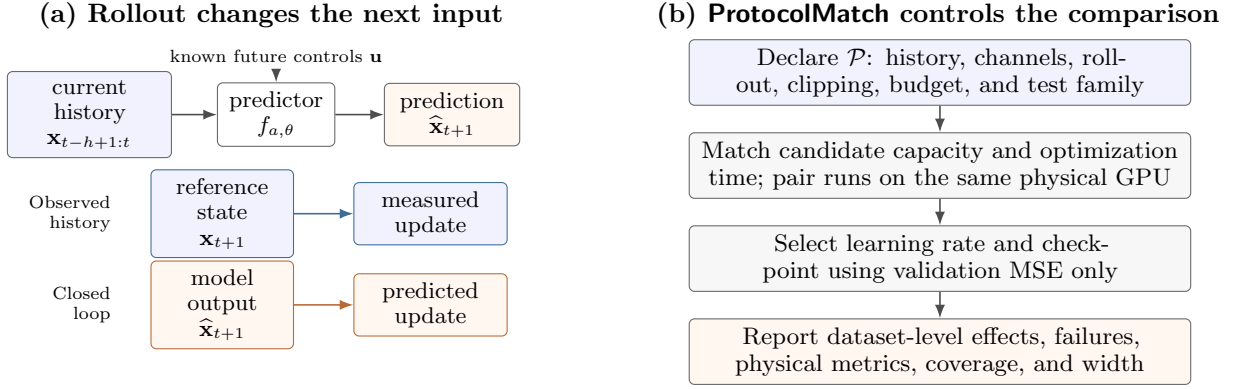
\begin{figure*}[t]
\centering
\begin{minipage}[t]{0.49\textwidth}
\centering
\textbf{(a) Rollout changes the next input}\par\smallskip
\begin{tikzpicture}[
  scale=0.95,
  transform shape,
  box/.style={draw=black!60,fill=white,rounded corners=2pt,align=center,minimum height=8mm,font=\footnotesize},
  arr/.style={-{Latex[length=1.8mm]},semithick,draw=black!70}
]
\path[use as bounding box] (-0.15,-3.10) rectangle (7.65,1.05);
\node[box,fill=blue!5,text width=2.05cm] (history) at (1.20,0) {current history\\\(\mathbf{x}_{t-h+1:t}\)};
\node[box,text width=1.40cm] (model) at (3.80,0) {predictor\\\(f_{a,\theta}\)};
\node[box,fill=orange!6,text width=1.65cm] (prediction) at (6.25,0) {prediction\\\(\widehat{\mathbf{x}}_{t+1}\)};
\node[font=\scriptsize] (controls) at (3.80,0.82) {known future controls \(\mathbf{u}\)};
\draw[arr] (history) -- (model);
\draw[arr] (controls) -- (model);
\draw[arr] (model) -- (prediction);

\node[font=\scriptsize,align=right,anchor=east] at (1.62,-1.38) {Observed\\history};
\node[box,draw=datasetblue,fill=blue!5,text width=1.75cm] (reference) at (3.05,-1.38) {reference state\\\(\mathbf{x}_{t+1}\)};
\node[box,draw=datasetblue,fill=blue!5,text width=1.90cm] (refresh-next) at (5.95,-1.38) {measured\\update};
\draw[arr,draw=datasetblue] (reference) -- (refresh-next);

\node[font=\scriptsize,align=right,anchor=east] at (1.62,-2.65) {Closed\\loop};
\node[box,draw=datasetorange,fill=orange!6,text width=1.75cm] (feedback) at (3.05,-2.65) {model output\\\(\widehat{\mathbf{x}}_{t+1}\)};
\node[box,draw=datasetorange,fill=orange!6,text width=1.90cm] (closed-next) at (5.95,-2.65) {predicted\\update};
\draw[arr,draw=datasetorange] (feedback) -- (closed-next);
\end{tikzpicture}
\end{minipage}\hfill
\begin{minipage}[t]{0.48\textwidth}
\centering
\textbf{(b) \method{} controls the comparison}\par\smallskip
\begin{tikzpicture}[
  node distance=3.5mm,
  flow/.style={draw=black!60,fill=white,rounded corners=2pt,align=center,minimum height=7mm,text width=6.4cm,font=\footnotesize},
  arr/.style={-{Latex[length=1.8mm]},semithick,draw=black!70}
]
\node[flow,fill=blue!5] (protocol) {Declare \(\protocol\): history, channels, rollout, clipping, budget, and test family};
\node[flow,fill=black!3,below=of protocol] (candidates) {Match candidate capacity and optimization time; pair runs on the same physical GPU};
\node[flow,fill=black!3,below=of candidates] (select) {Select learning rate and checkpoint using validation MSE only};
\node[flow,fill=orange!6,below=of select] (report) {Report dataset-level effects, failures, physical metrics, coverage, and width};
\draw[arr] (protocol) -- (candidates);
\draw[arr] (candidates) -- (select);
\draw[arr] (select) -- (report);
\end{tikzpicture}
\end{minipage}
\caption{The forecasting protocol determines both the information available to a predictor and how that predictor is compared. (a) Observed-history and closed-loop evaluation keep the predictor, future controls, and target fixed; only the source of the next state-history window changes. (b) \method{} pairs that declared task with controlled candidates, validation-only selection, independent-dataset aggregation, and failure-preserving reliability reports.}
\label{fig:framework}
\end{figure*}

\subsection{Forecasting task}

Let \(\mathbf{x}_t\in\mathbb{R}^{d}\) denote the vector of observed physical quantities at discrete time \(t\), and let \(\mathbf{u}_t\in\mathbb{R}^{q}\) denote known controls. Given a history of \(h\) states and controls through a future horizon of \(p\) steps, a forecaster \(f_{a,\theta}\) from architecture family \(a\in\archset\) produces
\begin{equation}
\widehat{\mathbf{x}}_{t+1:t+p}
=f_{a,\theta}\!\left(\mathbf{x}_{t-h+1:t},
\mathbf{u}_{t-h+1:t+p}\right),
\end{equation}
where \(\theta\) denotes trainable parameters. The controls are known throughout the forecast; only the state history may be restricted.

A forecasting protocol is the tuple
\begin{equation}
\protocol=(h,S,r,c,B,Q),
\end{equation}
where \(h\) is available history length, \(S\subseteq\{1,\ldots,d\}\) is the set of observed state channels, \(r\) is the rollout rule, \(c\) specifies output clipping, \(B\) is the optimization-time budget, and \(Q\) is the evaluation distribution. We consider two rollout rules. \emph{Observed-history} forecasting refreshes every input window with the reference trajectory. \emph{Closed-loop} forecasting feeds predictions into later windows. Figure~\ref{fig:framework}(a) makes the changed edge explicit: future controls and the prediction target remain fixed, but the source of the next history window changes. The two protocols therefore measure local block accuracy and autonomous stability, respectively.

Let \(j\) index forecast blocks; let \(\mathbf{H}_{j}^{(r)}\) denote the length-\(h\) state-history window under rule \(r\), \(\mathbf{U}_{j}\) the required known controls, and \(\mathbf{X}_{j}\) and \(\widehat{\mathbf{X}}_{j}\) the reference and predicted \(p\)-step state blocks. Let \(\mathbin{\Vert}\) denote concatenation and \(\operatorname{tail}_{h}\) retain the most recent \(h\) states. Then
\begin{equation}
\begin{aligned}
\widehat{\mathbf{X}}_{j}
  &= f_{a,\theta}\!\left(\mathbf{H}_{j}^{(r)},\mathbf{U}_{j}\right),\\
\mathbf{H}_{j+1}^{(r)}
  &=
  \begin{cases}
  \operatorname{tail}_{h}\!\left(\mathbf{H}_{j}^{(r)}\mathbin{\Vert}\mathbf{X}_{j}\right),
    & r=\mathrm{observed},\\
  \operatorname{tail}_{h}\!\left(\mathbf{H}_{j}^{(r)}\mathbin{\Vert}\widehat{\mathbf{X}}_{j}\right),
    & r=\mathrm{closed}.
  \end{cases}
\end{aligned}
\label{eq:rollout-update}
\end{equation}
Both rules start from the same history. Equation~\eqref{eq:rollout-update} changes only the conditioning states after the first block: reference states for observed-history evaluation and model outputs for closed-loop evaluation.

For a fixed protocol, the risk of architecture \(a\) is
\begin{equation}
\risk_{\protocol}(a)=
\mathbb{E}_{\dataset\sim Q}
\left[\mse\!\left(f_{a,\widehat\theta_{a}};\dataset,\protocol\right)\right],
\end{equation}
where \(\widehat\theta_a\) is selected without test data. A ranking reversal occurs when two protocols \(\protocol_1\) and \(\protocol_2\) satisfy
\(\risk_{\protocol_1}(a)<\risk_{\protocol_1}(b)\) but
\(\risk_{\protocol_2}(a)>\risk_{\protocol_2}(b)\).

\subsection{Comparative estimand}

For dataset realization \(d\), optimization seed \(s\), and protocol \(\protocol\), let \(e_{a,d,s,\protocol}\) be the test error of the validation-selected predictor from architecture \(a\). A paired architecture contrast is
\begin{equation}
\Delta_{d,s,\protocol}(a,b)
=e_{a,d,s,\protocol}-e_{b,d,s,\protocol}.
\end{equation}
For \(n_s\) optimization seeds per dataset and \(n_d\) independently generated datasets, the reported contrast is hierarchical:
\begin{equation}
\begin{aligned}
\overline{\Delta}_{d,\protocol}(a,b)
&=\frac{1}{n_s}\sum_{s=1}^{n_s}\Delta_{d,s,\protocol}(a,b),\\
\widehat{\Delta}_{\protocol}(a,b)
&=\frac{1}{n_d}\sum_{d=1}^{n_d}\overline{\Delta}_{d,\protocol}(a,b).
\end{aligned}
\label{eq:hierarchical-contrast}
\end{equation}
Each dataset-level sign identifies the better model under that protocol, while \(\widehat{\Delta}_{\protocol}\) summarizes independently regenerated datasets. Optimization seeds share trajectories; dataset seeds regenerate every split and trajectory.

The protocol is not a nuisance covariate to average away. If \(\Delta\) changes sign between observed-history and closed-loop rollout, the two risks correspond to different deployment behaviors. Our primary object is therefore the map
\begin{equation}
(a,b,\protocol)\longmapsto
\operatorname{sign}\!\left[\mathbb{E}_{d}\Delta_{d,\protocol}(a,b)\right],
\end{equation}
supplemented by effect magnitudes, failure counts, and reliability metrics. Universal architecture claims require this map to remain stable over the relevant protocol set; conditional recommendations do not.

\subsection{The \method{} evaluation framework}

Figure~\ref{fig:framework}(b) summarizes the comparison workflow. Four controls make the resulting rank interpretable.

\paragraph{Matched candidates.}
Model widths or ranks are chosen to target 150,000 trainable parameters within 10\%. Every architecture--learning-rate pair is trained on the same physical GPU within a comparison block. Checkpoints are recorded after 30 and 120 seconds of measured optimization time; data preparation, validation, and checkpoint I/O are outside this clock.

\paragraph{Validation-only selection.}
For each architecture, independent dataset, and training seed, the learning rate and checkpoint state are chosen by nonidentity-observable validation MSE. Test outcomes never determine selection or reruns.

\paragraph{Replication hierarchy.}
Three training seeds share each generated dataset and are averaged first. Three independently generated datasets are then the replication units for the core and history experiments. This prevents optimization seeds from being reported as independent physical datasets.

\paragraph{Failure preservation.}
Evaluation eligibility is fixed before test outcomes are examined. Nonfinite \mbox{predictions} count as failures; clipping defines a different protocol rather than repairing the original result. An infrastructure failure remains missing instead of being replaced by another seed.

\begin{table}[b]
\caption{Protocol factors varied in the evaluation grid.}
\label{tab:protocol-grid}
\centering
\small
\begin{tabular}{@{}l>{\raggedright\arraybackslash}p{1.55cm}>{\raggedright\arraybackslash}p{3.95cm}@{}}
\toprule
Factor & Values & Question \\
\midrule
History & 1, 5, 20 & How much past state is useful? \\
Channels & Full, local & Which observables are available? \\
Rollout & Refresh, closed loop & Is feedback autonomous? \\
Clipping & Off, on & Are bounds imposed at inference? \\
Budget & 30 s, 120 s & Is optimization time matched? \\
Test family & ID, freq., state shifts & Where must the model generalize? \\
\bottomrule
\end{tabular}
\end{table}

The framework is deliberately \emph{failure preserving}. A model that becomes nonfinite under one row of Table~\ref{tab:protocol-grid} is reported as such, and a clipped forecast is not used as the outcome of an unclipped condition. This keeps the comparison denominator fixed and prevents test success from becoming an implicit hyperparameter.

\section{Experimental Design: Quantum Dynamics Testbed}

\subsection{Systems, observations, and shifts}

We simulate open chains with \(\hbar=1\). For \(n\) spins, the driven transverse-field Ising Hamiltonian is
\begin{equation}
H_{\mathrm{I}}(t)=g\sum_{i=1}^{n}X_i+B(t)\sum_{i=1}^{n-1}Z_iZ_{i+1},
\end{equation}
where \(X_i\), \(Y_i\), and \(Z_i\) are Pauli operators on site \(i\), \(g\in[0.5,2.0]\) is constant within a trajectory, and \(B(t)\) is the time-varying control. The two-spin anisotropic Heisenberg (XXZ) variant uses
\begin{align}
H_{\mathrm{H}}(t)&=g\sum_iX_i+B(t)\sum_i \bigl(
X_iX_{i+1}+Y_iY_{i+1} \notag\\
&\hspace{35mm}+0.7Z_iZ_{i+1}\bigr).
\end{align}

Each trajectory contains 401 points with step \(\Delta t=0.02\), spanning \(t\in[0,8]\); supervised targets used for training end at \(t=4\). Training controls cycle through sine, spline, quench, and Gaussian-process families. The frequency-shift split uses sine frequencies in \([1.5,2.0]\), outside the training sine range \([0.1,1.0]\). Two additional test families replace random product initial states by aligned or entangled states. The identity-distribution test retains the training families with disjoint trajectories.

For two and four spins, \(\mathbf{x}_t\) contains all \(4^n\) Pauli expectations. The six-spin extension predicts 19 quantities: identity and the three single-site Pauli expectations at each site. It is a local-observable forecasting task, not full-state reconstruction; its absolute error is therefore not directly comparable with the full-Pauli tasks.

\subsection{Models and optimization}

We compare four families: (i) an LSTM forecaster, (ii) a PatchTST-style model with effective LayerNorm, (iii) a causally masked Transformer, and (iv) a rank-constrained linear predictor. Every model consumes the same known controls and predicts a residual relative to the latest visible state. The default history and forecast lengths are both 20 steps. For partial-input experiments, unobserved history entries are set to zero while all target channels remain evaluated.

Two diagnostic baselines test whether learned state history is needed at all. A current-state MLP uses only the latest full observation and the complete known control sequence, passes them through one GELU hidden layer, and predicts a 20-step residual. A parameter-free persistence rule repeats the latest available observation. We analyze these controls separately from the four-family comparison because persistence has no fitting stage and the MLP changes the information set as well as the architecture.

Models use AdamW with learning rates \(3\times10^{-4}\) or \(10^{-3}\), weight decay \(10^{-4}\), batch size 64, dropout 0.1 for neural models, gradient-norm clipping at 1, and no output clipping during training. Each dataset contains 256 training, 64 validation, 64 calibration, and 64 trajectories in each test family; the small-data core cell uses the first 64 training trajectories. Appendix~\ref{app:design} specifies the full factorial design and selection procedure.

For constraint experiments, the training objective is
\begin{equation}
\mathcal{L}(\widehat{\mathbf{x}},\mathbf{x})=
\mse(\widehat{\mathbf{x}},\mathbf{x})+
\lambda\sum_{v\in V}\mathcal{V}_{v}(\widehat{\mathbf{x}}),
\end{equation}
where \(\lambda\in\{0.001,0.01,0.1\}\) is the constraint weight and \(V\) is a selected subset of trace, positivity, and bound terms. For predicted Pauli coefficients \(\widehat{\mathbf{x}}\), trace violation penalizes \((\widehat{x}_{I}-1)^2\). Positivity reconstructs \(\widehat\rho=2^{-n}\sum_P\widehat{x}_{P}P\) and penalizes squared negative eigenvalue parts. The bound term penalizes squared excess beyond \([-1,1]\) for every observable. All terms are computed on full-Pauli predictions.

\subsection{Evaluation axes}

Prediction error is the mean squared error over nonidentity channels in either the same-time region \(t\leq4\) or the extrapolation region \(t>4\). Four test families, two rollout rules, clipping on or off, and two time regions yield 32 correlated views per condition. We report descriptive counts across these views and direction agreement across independent datasets; the 32 views are not treated as independent replicates.

Physical penalties target trace normalization, density-matrix positivity, and Pauli-observable bounds. Their diagnostic metrics are reported separately from prediction MSE. For uncertainty, each model uses 64 independent identity-distribution calibration trajectories. The score for one trajectory is its largest absolute nonidentity-channel error in the reported region. The 59th ordered score defines a symmetric radius targeting 90\% whole-trajectory-region coverage.

\begin{figure*}[t]
\centering
\begin{minipage}[t]{0.485\textwidth}
\centering
\begin{tikzpicture}[font=\footnotesize]
\path[use as bounding box] (-0.04,0) rectangle (8.24,6.0);
\node[font=\small] at (4.45,5.75) {(a) Architecture rank reverses with data};
\node[font=\small,rotate=90] at (0.12,3.2) {MSE difference: LSTM -- causal};
\node[font=\small] at (4.45,0.32) {Training trajectories};
\begin{scope}[shift={(2.15cm,3.62cm)},x=4.6cm,y=70cm]
\foreach \y/\value in {-0.03/{-0.03},-0.02/{-0.02},-0.01/{-0.01},0/0,0.01/0.01,0.02/0.02} {
  \draw[black!10] (-0.18,\y) -- (1.18,\y);
  \draw[black!65] (-0.18,\y) -- ++(-0.016,0);
  \node[anchor=east,xshift=-3pt] at (-0.18,\y) {\(\value\)};
}
\draw[black!65] (-0.18,0.024) -- (-0.18,-0.036) -- (1.18,-0.036);
\foreach \x/\value in {0/64,1/256} {
  \draw[black!65] (\x,-0.036) -- ++(0,-0.001);
  \node[anchor=north,yshift=-4pt] at (\x,-0.036) {\value};
}
\draw[black!55,densely dotted] (-0.18,0) -- (1.18,0);
\draw[black,very thick,dashed] plot coordinates {(0,0.0072101934264455015) (1,-0.02198415196631832)};
\draw[datasetblue,semithick,xshift=-4.5pt] plot[mark=*,mark size=2pt] coordinates {(0,0.012921963250798548) (1,-0.026960506423744337)};
\draw[datasetorange,semithick,dashed] plot[mark=square*,mark size=2pt] coordinates {(0,0.005076709094874093) (1,-0.015856072864134405)};
\draw[datasetgreen,semithick,dashdotted,xshift=4.5pt] plot[mark=triangle*,mark size=2.3pt] coordinates {(0,0.003631907933663863) (1,-0.02313587661107622)};
\node[text=black!65,anchor=west] at (-0.12,0.019) {Causal better};
\node[text=black!65,anchor=west] at (-0.12,-0.032) {LSTM better};
\end{scope}
\end{tikzpicture}
\end{minipage}\hfill
\begin{minipage}[t]{0.485\textwidth}
\centering
\begin{tikzpicture}[font=\footnotesize]
\path[use as bounding box] (-0.04,0) rectangle (8.24,6.0);
\node[font=\small] at (4.45,5.75) {(b) Input rank reverses with rollout};
\node[font=\small,rotate=90] at (0.12,3.2) {MSE difference: local -- full};
\node[font=\small] at (4.45,0.32) {Rollout rule};
\begin{scope}[shift={(2.15cm,3.4cm)},x=4.6cm,y=50cm]
\foreach \y/\value in {-0.04/{-0.04},-0.02/{-0.02},0/0,0.02/0.02} {
  \draw[black!10] (-0.18,\y) -- (1.18,\y);
  \draw[black!65] (-0.18,\y) -- ++(-0.016,0);
  \node[anchor=east,xshift=-3pt] at (-0.18,\y) {\(\value\)};
}
\draw[black!65] (-0.18,0.038) -- (-0.18,-0.046) -- (1.18,-0.046);
\foreach \x/\value in {0/Observed-history,1/Closed-loop} {
  \draw[black!65] (\x,-0.046) -- ++(0,-0.0015);
  \node[anchor=north,yshift=-4pt] at (\x,-0.046) {\value};
}
\draw[black!55,densely dotted] (-0.18,0) -- (1.18,0);
\draw[black,very thick,dashed] plot coordinates {(0,0.020464528049018018) (1,-0.03022718506736897)};
\draw[datasetblue,semithick,xshift=-4.5pt] plot[mark=*,mark size=2pt] coordinates {(0,0.019666941926260728) (1,-0.032210111620385076)};
\draw[datasetorange,semithick,dashed] plot[mark=square*,mark size=2pt] coordinates {(0,0.020106457096783034) (1,-0.03071596186288539)};
\draw[datasetgreen,semithick,dashdotted,xshift=4.5pt] plot[mark=triangle*,mark size=2.3pt] coordinates {(0,0.021620185124010295) (1,-0.02775548171883645)};
\node[text=black!65,anchor=west] at (-0.12,0.031) {Full history better};
\node[text=black!65,anchor=west] at (-0.12,-0.041) {Local history better};
\end{scope}
\end{tikzpicture}
\end{minipage}
\par\smallskip
\begin{tikzpicture}[font=\footnotesize]
\draw[datasetblue,thin] (0,0) -- (0.55,0);
\fill[datasetblue] (0.275,0) circle (2pt);
\node[anchor=west] at (0.65,0) {Dataset 1};
\draw[datasetorange,thin,dashed] (2.8,0) -- (3.35,0);
\node[fill=datasetorange,minimum size=4pt,inner sep=0pt] at (3.075,0) {};
\node[anchor=west] at (3.45,0) {Dataset 2};
\draw[datasetgreen,thin,dashdotted] (5.6,0) -- (6.15,0);
\node[text=datasetgreen] at (5.875,0) {\(\blacktriangle\)};
\node[anchor=west] at (6.25,0) {Dataset 3};
\draw[black,thick,dashed] (8.4,0) -- (8.95,0);
\node[anchor=west] at (9.05,0) {Dataset mean};
\end{tikzpicture}
\caption{Replicated ranking reversals under matched optimization time. (a) LSTM-minus-causal error for closed-loop extrapolation changes sign between 64 and 256 training trajectories. (b) Local-minus-full-history error changes sign between observed-history and closed-loop rollout while the full prediction target remains fixed. Colored lines are independent-dataset means over three training seeds; the dashed black line is their average. Both panels use 20-step histories, a 120-second budget, identity-distribution tests, and no output clipping.}
\label{fig:reversals}
\end{figure*}
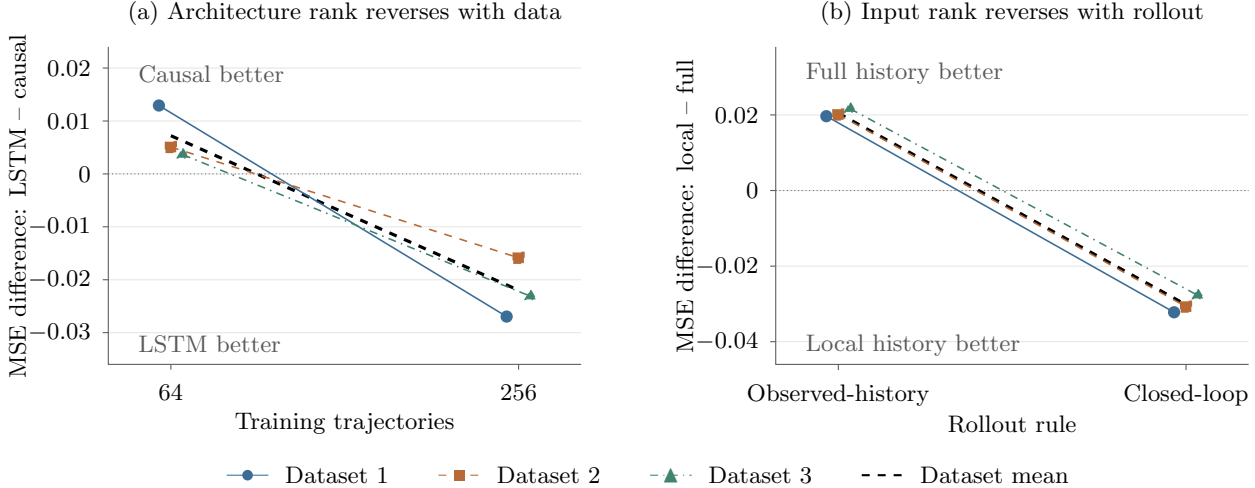

\section{Results}

We organize the evidence around four prespecified questions. Which predictor is preferred as data amount and target space change? Does the value of an information channel survive a change in rollout? Do physical constraints improve the metric used for model selection? Does held-out calibration remain reliable under a known distribution shift? Each subsection changes one part of the protocol while holding the comparison target fixed; exact baseline and input contrasts appear in Appendices~\ref{app:baselines} and~\ref{app:input-effects}.

\subsection{Data regime reverses the architecture ranking}

Does the preferred architecture change when the same forecasting task receives more training data? Figure~\ref{fig:reversals}(a) compares the LSTM and causal Transformer on two-spin Ising dynamics with 64 or 256 training trajectories; the smaller set is a fixed prefix of the larger one. The target, available history, and evaluation protocol are unchanged. This prespecified high-stress slice uses a 120-second optimization budget, identity-distribution test trajectories, unclipped closed-loop rollout, and extrapolation beyond \(t=4\).

The LSTM-minus-causal MSE changes from \(+0.00721\) with 64 trajectories to \(-0.02198\) with 256. Each of the three independent dataset contrasts crosses zero in the same direction. Thus the architecture selected for lower error reverses from causal attention to recurrence: additional data changes the ordering, not merely the size of a performance gap.

Table~\ref{tab:core} places this comparison alongside the other system and target cells. Each mean averages training seeds within datasets before averaging datasets.

\begin{table*}[t]
\caption{Nonidentity-observable MSE under unclipped closed-loop extrapolation. Bold marks the lowest mean within each row. Two/four-spin targets are full Pauli vectors; the six-spin target contains local observables only.}
\label{tab:core}
\centering
\small
\begin{tabular}{lrrrrr}
\toprule
System and target & Train traj. & LSTM & PatchTST & Causal Transformer & Low-rank linear \\
\midrule
Two-spin Ising, full Pauli & 64  & 0.09402 & 0.65869 & \winner{0.08681} & 0.19583 \\
Two-spin Ising, full Pauli & 256 & \winner{0.02456} & 0.66745 & 0.04654 & 0.18849 \\
Four-spin Ising, full Pauli & 256 & \winner{0.10608} & 0.12416 & 0.10902 & 0.12220 \\
Two-spin XXZ, full Pauli & 256 & \winner{0.03074} & 0.25047 & 0.04540 & 0.18421 \\
Six-spin Ising, local Pauli & 256 & 0.08833 & 0.08835 & 0.07878 & \winner{0.07668} \\
\bottomrule
\end{tabular}
\end{table*}

The LSTM also leads on the two-spin XXZ and four-spin full-observable cells. On the six-spin local-observable task, the low-rank linear predictor attains the lowest mean, followed by causal attention. For every displayed LSTM--causal comparison, all three dataset-level means have the same sign. The target-space comparison complements the within-task reversal in Figure~\ref{fig:reversals}(a): different observation targets favor different model families, while a fixed full-Pauli target already suffices to produce a reversal across data regimes.

The two attention models have distinct outcomes. PatchTST has higher error in every row, whereas causal attention leads in the small-data cell and is near the best model in the four-spin cell. Including both, together with a linear control, makes the model-selection conclusion specific: the preferred sequence prior depends on data amount and the prediction target.

\subsection{State refresh changes the simple-baseline ranking}

Can a predictor that sees only the latest state replace a temporal model? Table~\ref{tab:baselines} compares the current-state MLP with persistence under the same identity-distribution, 120-second, extrapolation protocol used for the main high-stress comparisons. The forecast differs only in whether it uses a learned latest-state map or an unchanged-state rule; the rollout protocol then determines how the next input state is supplied.

\begin{table*}[t]
\caption{Latest-state controls on identity-distribution extrapolation. Entries are nonidentity MSE mean $\pm$ sample SD across three independent dataset means; MLP training seeds are averaged within each dataset first. The MLP uses its validation-selected 120-second checkpoint. ``3/3'' denotes the same paired direction in all datasets.}
\label{tab:baselines}
\centering
\scriptsize
\setlength{\tabcolsep}{2.5pt}
\begin{tabular}{@{}lrcccl@{}}
\toprule
& & \multicolumn{2}{c}{Observed-history refresh} & \multicolumn{2}{c}{Unclipped closed loop} \\
\cmidrule(lr){3-4}\cmidrule(lr){5-6}
System and target & Train traj. & Current-state MLP & Persistence & Current-state MLP & Persistence / direction \\
\midrule
Two-spin Ising, full Pauli & 64  & $0.00524\pm0.00094$ & $0.09979\pm0.00323$ & $0.49307\pm0.31782$ & $0.34285\pm0.00661$ / mixed \\
Two-spin Ising, full Pauli & 256 & $0.00200\pm0.00038$ & $0.09979\pm0.00323$ & $0.22436\pm0.03245$ & $0.34285\pm0.00661$ / MLP 3/3 \\
Four-spin Ising, full Pauli & 256 & $0.03675\pm0.00150$ & $0.04238\pm0.00158$ & $2.73{\times}10^{35}\pm4.72{\times}10^{35}$ & $0.11203\pm0.00119$ / mixed \\
Two-spin XXZ, full Pauli & 256 & $0.00630\pm0.00306$ & $0.12930\pm0.00878$ & $4.12985\pm5.75518$ & $0.33538\pm0.00844$ / persistence 3/3 \\
Six-spin Ising, local Pauli & 256 & $0.01362\pm0.00148$ & $0.03222\pm0.00299$ & $0.08737\pm0.00167$ & $0.29335\pm0.00579$ / MLP 3/3 \\
\bottomrule
\end{tabular}
\end{table*}

With observed-history refresh, the MLP has lower error on every dataset in all five cells. Under closed-loop feedback, it remains lower in the two-spin 256-trajectory and six-spin local-observable cells, persistence is lower on all three XXZ datasets, and the remaining two cells are mixed. The four-spin mean includes finite explosive predictions rather than replacing them with stable fits. Output clipping changes the comparison again: the MLP is lower on all three datasets in four of five cells, while the four-spin direction remains mixed. A latest-state map can therefore be an accurate refreshed-state predictor without being a stable autonomous simulator. Appendices~\ref{app:baseline-design} and~\ref{app:baselines} give the design and individual high-error values.

\subsection{Rollout reverses the value of nonlocal history}

Can restricting the observed state improve a forecast without simplifying its target? The four-spin experiment compares models trained with full history against models trained with only 13 local channels--identity plus all single-site Pauli expectations. Both predict the same full 256-channel vector and receive the same future controls. Nonlocal history entries are masked to zero; target channels are neither removed nor downweighted.

Figure~\ref{fig:reversals}(b) defines the input contrast as local-minus-full MSE. For the history-20 LSTM, it is \(+0.02046\) with observed-history refresh and \(-0.03023\) with closed-loop feedback. Each independent dataset shows the same sign change. The benefit of nonlocal history therefore depends on whether those channels are refreshed from observations or populated by preceding predictions.

This reversal extends across the evaluation grid. Local-only input worsens all 16 observed-history views and improves all 16 closed-loop views for each of the three neural architectures, every history length in \(\{1,5,20\}\), and all three datasets. These views cover test families, clipping choices, and time regions; their direction counts summarize shared-model evaluations rather than additional independent replications.

History length is conditional as well. With full inputs, LSTM history 5 has lower mean error than history 20 in 27 of 32 two-spin views and 30 of 32 four-spin views. All three dataset-level effects agree on the improvement in 9 and 28 views, respectively (Appendix~\ref{app:input-effects}). More available context does not imply a better forecast.

The operational consequence is direct: an information set should be selected under the rollout used at deployment. Observed-history refresh measures the value of additional reference channels; closed-loop evaluation also measures the consequences of feeding their predictions back. The measured sign reversal establishes this distinction. Error amplification through nonlocal feedback is a mechanistic hypothesis for the interaction, not a prerequisite for the model-selection result.

\subsection{Physical validity and accuracy are separate objectives}

Do constraints improve forecasting accuracy, or only the physical property they encode? On one \mbox{prespecified} dataset, a trace-normalization penalty improves its targeted trace metric in 31/32, 32/32, and 32/32 views at weights 0.001, 0.01, and 0.1 (Figure~\ref{fig:reliability}(a)). Prediction MSE simultaneously increases in 26/32, 17/32, and 21/32 views. Combining trace, positivity, and observable-bound penalties increases MSE in all 32 views for both the LSTM and causal Transformer at every tested weight (Appendix~\ref{app:constraint-effects}).

These are not contradictory outcomes: the loss defines a multi-objective optimization problem. Under a fixed wall-clock budget, constraint computation also consumes optimization time. Physical penalties should therefore be selected using the physical metric they are meant to improve and an explicit accuracy tolerance, rather than being assumed to regularize MSE monotonically.

\subsection{In-distribution calibration does not transfer to shift}

\begin{figure*}[t]
\centering
\begin{minipage}[t]{0.485\textwidth}
\centering
\begin{tikzpicture}[font=\footnotesize]
\path[use as bounding box] (-0.05,0) rectangle (8.2,5.25);
\node[font=\small] at (4.45,5.02) {(a) Validity and accuracy move separately};
\node[font=\small,rotate=90] at (0.12,2.78) {Views out of 32};
\node[font=\small] at (4.45,0.18) {Trace-penalty weight};
\begin{scope}[shift={(1.55cm,1.02cm)},x=2.15cm,y=0.10cm]
\foreach \y/\value in {0/0,8/8,16/16,24/24,32/32} {
  \draw[black!10] (-0.35,\y) -- (2.45,\y);
  \draw[black!65] (-0.35,\y) -- ++(-0.035,0);
  \node[anchor=east,xshift=-3pt] at (-0.35,\y) {\value};
}
\draw[black!65] (-0.35,34) -- (-0.35,0) -- (2.45,0);
\foreach \x/\value in {0/0.001,1/0.01,2/0.1} {
  \draw[black!65] (\x,0) -- ++(0,-0.8);
  \node[anchor=north,yshift=-3pt] at (\x,0) {\value};
}
\draw[datasetgreen,thick] plot[mark=triangle*,mark size=2.5pt] coordinates {(0,31) (1,32) (2,32)};
\draw[datasetorange,thick,dashed] plot[mark=square*,mark size=2.2pt] coordinates {(0,26) (1,17) (2,21)};
\end{scope}
\draw[datasetgreen,thick] (1.35,4.52) -- (1.9,4.52);
\node[text=datasetgreen] at (1.625,4.52) {\(\blacktriangle\)};
\node[anchor=west] at (2.0,4.52) {lower trace error};
\draw[datasetorange,thick,dashed] (4.65,4.52) -- (5.2,4.52);
\node[fill=datasetorange,minimum size=4pt,inner sep=0pt] at (4.925,4.52) {};
\node[anchor=west] at (5.3,4.52) {higher prediction MSE};
\end{tikzpicture}
\end{minipage}\hfill
\begin{minipage}[t]{0.485\textwidth}
\centering
\begin{tikzpicture}[font=\footnotesize]
\path[use as bounding box] (-0.05,0) rectangle (8.2,5.25);
\node[font=\small] at (4.45,5.02) {(b) Coverage degrades under frequency shift};
\node[font=\small,rotate=90] at (0.18,2.78) {Trajectory coverage (\%)};
\begin{scope}[shift={(1.70cm,1.18cm)},x=1.55cm,y=0.0305cm]
\foreach \y/\value in {0/0,25/25,50/50,75/75,100/100} {
  \draw[black!10] (-0.35,\y) -- (3.55,\y);
  \draw[black!65] (-0.35,\y) -- ++(-0.045,0);
  \node[anchor=east,xshift=-3pt] at (-0.35,\y) {\value};
}
\draw[black!65] (-0.35,100) -- (-0.35,0) -- (3.55,0);
\draw[black!55,densely dotted] (-0.35,90) -- (3.55,90);
\node[anchor=west,text=black!60,font=\scriptsize,fill=white,inner sep=1pt] at (0.00,96) {90\% target};
\foreach \x/\name/\width in {0/LSTM/1.28,1/Patch/7.65,2/Causal/1.72,3/Linear/2.33} {
  \draw[black!65] (\x,0) -- ++(0,-2.3);
  \node[anchor=north,align=center,yshift=-3pt] at (\x,0) {\name\\{\scriptsize width \width}};
}
\draw[datasetblue,thick] plot[mark=*,mark size=2.4pt] coordinates {(0,87.50) (1,91.32) (2,86.98) (3,87.67)};
\draw[datasetorange,thick,dashed] plot[mark=square*,mark size=2.2pt] coordinates {(0,28.13) (1,26.91) (2,16.49) (3,24.31)};
\end{scope}
\draw[datasetblue,thick] (2.05,4.52) -- (2.55,4.52);
\fill[datasetblue] (2.30,4.52) circle (2pt);
\node[anchor=west] at (2.65,4.52) {unshifted};
\draw[datasetorange,thick,dashed] (4.75,4.52) -- (5.25,4.52);
\node[fill=datasetorange,minimum size=4pt,inner sep=0pt] at (5.00,4.52) {};
\node[anchor=west] at (5.35,4.52) {frequency shift};
\end{tikzpicture}
\end{minipage}
\caption{Accuracy, physical validity, and uncertainty are distinct outcomes. (a) A trace penalty usually lowers trace error but often raises prediction MSE for a two-spin LSTM; the 32 correlated views share fitted models. (b) Intervals calibrated on 64 independent identity-distribution trajectories approach nominal 90\% coverage in distribution but lose most coverage after a frequency shift. Widths are full symmetric interval widths.}
\label{fig:reliability}
\end{figure*}
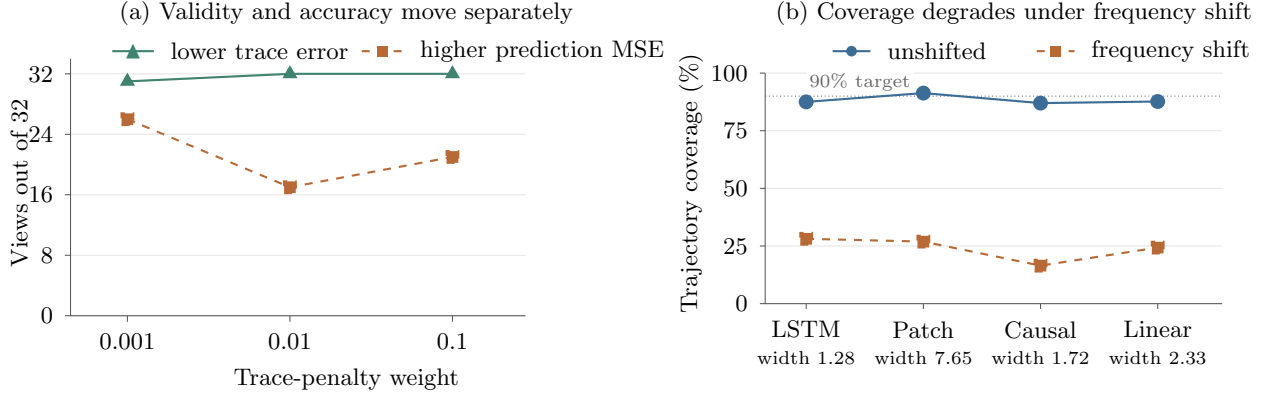

Does held-out calibration protect a forecaster under shift? Figure~\ref{fig:reliability}(b) fixes the two-spin Ising, 256-trajectory, full-history condition from Table~\ref{tab:core}. Coverage requires every nonidentity prediction in the extrapolation region of a trajectory to lie inside the interval. All models are calibrated only on independent identity-distribution trajectories.

Identity-distribution coverage is 87--91\%, but frequency-shift coverage falls to 16--28\%. PatchTST attains the highest identity coverage only with intervals roughly six times wider than the LSTM's. Coverage, sharpness, and shifted reliability must therefore be evaluated together; a fixed identity-calibrated radius is not a shift detector. Exact values and the calibration construction appear in Appendix~\ref{app:calibration}.

\subsection{Failure-preserving accounting bounds the claim}

What changes when unsuccessful evaluations remain visible? The design defines 5,616 protocol--region intervals; 5,608 are available, while eight remain missing because one validation-selected predictor could not initialize on the evaluation device. Unclipped feedback also exposes a nonfinite PatchTST forecast, whereas clipping makes the forecast finite but substantially less accurate. The latest-state baseline retains finite explosive four-spin outcomes and one high-error XXZ seed. None is replaced by a successful run, so test-conditioned reruns cannot silently change the estimand (Appendix~\ref{app:failures}).

A paired six-fit PatchTST diagnostic compares BatchNorm with LayerNorm under shared initialization and matched hardware; the validation difference changes sign across seeds and budgets. Because the primary comparisons use effective LayerNorm, PatchTST underperformance is a result for the evaluated models and protocols, not evidence that one normalization layer or attention as a class is inherently unsuitable.

\section{Discussion}

The selected object is the pair \((a,\protocol)\), not an architecture name alone. Recurrence has the lowest error in the data-rich two-spin full-observable cell, causal attention leads in the smaller-data cell, and the linear predictor leads for six-spin local observables. The latest-state MLP is uniformly preferable to persistence when reference states refresh its input, yet its closed-loop ranking is system dependent. This conditional view extends beyond quantum dynamics: weather, control, molecular simulation, and learned surrogates likewise vary in available measurements, rollout feedback, compute, and distribution shift. A benchmark that changes architecture while averaging across these conditions does not identify the predictor appropriate for deployment.

Our evidence comes from exact simulated closed-system spin dynamics, up to four spins with complete Pauli observations and six spins with local observations. It does not estimate rankings for large many-body systems, noisy measurements, or open-system dynamics, and the constraint study uses one independent dataset rather than three.

The reversals motivate testable mechanisms but do not establish them. With complete Pauli inputs, a recurrent hidden state may compress the trajectory efficiently once enough training data are available. Under closed-loop rollout, every predicted channel can re-enter later windows, so nonlocal channels add feedback paths absent from local-only input. The six-spin linear result may instead reflect its local target rather than system size. Pauli-weight noise interventions and multi-site six-spin targets would distinguish these explanations.

The practical recommendation is to declare the protocol grid, pair candidate runs on hardware, select with validation data only, aggregate independently generated datasets, and retain failures. Accuracy, physical validity, and uncertainty should remain separate coordinates. Observed-history validation is insufficient for autonomous deployment, just as identity-distribution coverage is insufficient when shifts are expected. The resulting rule is conditional but actionable: select the predictor with the lowest validated risk under the intended observation, rollout, compute, and distributional conditions.

\section{Conclusion}

Scientific dynamics forecasting does not have a context-free architecture ranking. In controlled quantum-dynamics experiments, recurrent, attention-based, and linear predictors exchange advantages across data regimes and target spaces, while both the value of observed history and the stability of a latest-state MLP change with rollout. Physical penalties improve targeted consistency without reliably improving MSE, and identity-calibrated intervals lose coverage under frequency shift. By making protocol part of the model-selection estimand, the proposed \method{} converts these conditional outcomes into explicit selection and reporting rules for scientific forecasters. More broadly, this protocol-centered view turns architecture comparison from a search for a universal winner into a reproducible procedure for selecting models under the conditions in which they will actually be used.

\label{page:lastmain}

\section*{AI Use Statement}

Generative AI tools were used for language editing, manuscript-structure suggestions, and implementation assistance. The authors take responsibility for the final content of this work.

\bibliography{references}

\section*{Checklist}

\begin{enumerate}

  \item For all models and algorithms presented, check if you include:
  \begin{enumerate}
    \item A clear description of the mathematical setting, assumptions, algorithm, and/or model. [Yes]
    \item An analysis of the properties and complexity (time, space, sample size) of any algorithm. [Yes]
    \item (Optional) Anonymized source code, with specification of all dependencies, including external libraries. [No]
  \end{enumerate}

  \item For any theoretical claim, check if you include:
  \begin{enumerate}
    \item Statements of the full set of assumptions of all theoretical results. [Not Applicable]
    \item Complete proofs of all theoretical results. [Not Applicable]
    \item Clear explanations of any assumptions. [Yes]
  \end{enumerate}

  \item For all figures and tables that present empirical results, check if you include:
  \begin{enumerate}
    \item The code, data, and instructions needed to reproduce the main experimental results (either in the supplemental material or as a URL). [No]
    \item All the training details (e.g., data splits, hyperparameters, how they were chosen). [Yes]
    \item A clear definition of the specific measure or statistics and error bars (e.g., with respect to the random seed after running experiments multiple times). [Yes]
    \item A description of the computing infrastructure used. (e.g., type of GPUs, internal cluster, or cloud provider). [Yes]
  \end{enumerate}

  \item If you are using existing assets (e.g., code, data, models) or curating/releasing new assets, check if you include:
  \begin{enumerate}
    \item Citations of the creator If your work uses existing assets. [No]
    \item The license information of the assets, if applicable. [No]
    \item New assets either in the supplemental material or as a URL, if applicable. [No]
    \item Information about consent from data providers/curators. [Not Applicable]
    \item Discussion of sensible content if applicable, e.g., personally identifiable information or offensive content. [Not Applicable]
  \end{enumerate}

  \item If you used crowdsourcing or conducted research with human subjects, check if you include:
  \begin{enumerate}
    \item The full text of instructions given to participants and screenshots. [Not Applicable]
    \item Descriptions of potential participant risks, with links to Institutional Review Board (IRB) approvals if applicable. [Not Applicable]
    \item The estimated hourly wage paid to participants and the total amount spent on participant compensation. [Not Applicable]
  \end{enumerate}

\end{enumerate}

\clearpage
\appendix
\thispagestyle{empty}
\onecolumn
\preprinttitle{Protocol-Dependent Model Selection for Scientific Dynamics Forecasting:\\Appendix}

\section{Complete Experimental Design}
\label{app:design}

\subsection{Core matrix}

The core matrix crosses four architecture families with the following system/data cells: two-spin Ising with 64 or 256 training trajectories, four-spin Ising with 256 trajectories, and two-spin XXZ with 256 trajectories. For each cell, three independently generated dataset suites, three training seeds, two learning rates, and two optimization budgets are evaluated. The bounded six-spin extension repeats the same four architectures with 256 training trajectories and local-observable targets. These groups contain 288 core fits and 72 six-spin fits.

History and partial-input controls add 540 fits across two- and four-spin Ising systems, three neural architectures, available histories of 1, 5, or 20 steps, and full or local input. Constraint controls add 216 fits on a prespecified two-spin dataset across three architectures, four constraint sets, three weights, three training seeds, and two learning rates. The primary evaluation matrix contains 1,116 fitted candidates. A separate baseline study adds 90 current-state MLP fits and 15 deterministic persistence evaluations.

\begin{table}[h]
\caption{Experimental inventory. Counts denote fitted architecture/configuration instances except for the diagnostic row, which includes 90 MLP fits and 15 deterministic persistence evaluations.}
\label{tab:inventory}
\centering
\begin{tabular}{lrrl}
\toprule
Group & Evaluations & Independent datasets & Primary purpose \\
\midrule
Core matched comparisons & 288 & 3 per system cell & Architecture and data-regime ranks \\
History and input controls & 540 & 3 per condition & Information--rollout interaction \\
Constraint terms and weights & 216 & 1 & Accuracy--validity trade-off \\
Six-spin extension & 72 & 3 & Local-observable scaling check \\
\midrule
Primary matrix total & 1,116 & --- & --- \\
Diagnostic baselines (separate) & 105 & 3 per system cell & Latest-state and persistence controls \\
\bottomrule
\end{tabular}
\end{table}

\subsection{Data generation}

Each dataset suite uses an independent root seed and disjoint random-number substreams for split, trajectory, state, coupling, driving field, and trajectory identifier. No trajectory is skipped, repaired, clipped, or renormalized. Each suite contains train, validation, calibration, identity test, frequency-shift test, aligned-state test, and entangled-state test splits. Counts are 256/64/64/64 per test family, except that the small-data core condition uses a fixed 64-trajectory prefix of its training split.

Reference evolution uses complex128 propagation with left-endpoint float32 controls. Two- and four-spin systems store the lexicographically ordered full Pauli basis; six-spin systems store identity followed by site-major \(X\), \(Y\), and \(Z\) observables. Stored reference expectations are finite, real, normalized on identity, and within Pauli bounds.

\subsection{Training-time matching and selection}

The optimization clock includes device transfers, forward and backward passes, optimizer updates, gradient clipping, and finite-value checks. It excludes data loading before training, validation, checkpoint serialization, and result aggregation. A checkpoint is emitted at the first completed update after each target budget. Validation is performed on all candidate learning-rate/checkpoint pairs, and the lowest nonidentity MSE selects the model before test evaluation.

Within every training-seed comparison block, the four architectures use the same physical GPU. Hardware differs across blocks, so the design supports within-block matched-time comparisons but not a global throughput ranking across GPU models. Experiments use NVIDIA RTX A5000, RTX A6000, RTX 3090, and TITAN RTX devices.

\subsection{Diagnostic baseline design}
\label{app:baseline-design}

The current-state MLP concatenates the latest full observation with the known control sequence over the 20-step history and 20-step forecast, applies one GELU hidden layer with dropout 0.1, and predicts a residual from the latest observation. Its width is selected to target 150,000 trainable parameters within 10\%. The persistence rule has no trained parameters and repeats the latest available observation for all 20 forecast steps.

The baseline study crosses five system/data cells, three independent datasets, three MLP training seeds, two learning rates, and synchronized 30- and 120-second checkpoints. This produces 90 MLP fits; learning rate is selected separately for each seed and budget using same-time validation MSE only. Persistence is evaluated once for each system/data/dataset cell, producing 15 evaluations. All 105 planned evaluations completed, and no test outcome enters selection.

\section{Reporting and Replication Units}

Let \(e_{d,s}\) denote a scalar evaluation for dataset seed \(d\) and training seed \(s\). We first compute \(\bar e_d=\frac{1}{3}\sum_s e_{d,s}\), then report \(\frac{1}{3}\sum_d\bar e_d\). Architecture differences are paired within \((d,s)\) on the same physical GPU. Statements that all datasets agree refer to the signs of the three \(\bar e_d\) differences. The 32 views within a condition share fitted models and test construction and are not independent replications.

All primary errors exclude the identity observable. The same-time region contains targets at or before the training cutoff \(t=4\); the extrapolation region contains targets after that cutoff. Observed-history evaluation uses the reference state for every input window. Closed-loop evaluation uses the model's preceding predictions once the initial context has been consumed. Clipped and unclipped predictions are separate protocols.

\section{Additional Result Detail}

\subsection{Latest-state and persistence controls}
\label{app:baselines}

Table~\ref{tab:baselines} reports the primary 120-second comparison. With true-state refresh, every one of the 15 dataset-level pairs favors the current-state MLP. With unclipped closed-loop feedback, the two-spin Ising 256-trajectory and six-spin local-observable cells favor the MLP in all datasets, the two-spin XXZ cell favors persistence in all datasets, and the two-spin Ising 64-trajectory and four-spin cells have mixed signs. With output clipping, four cells favor the MLP in all datasets and the four-spin cell remains mixed.

The four-spin closed-loop mean is dominated by four finite divergences in two of the three independent datasets, with seed-level MSEs of $9.90\times10^{32}$, $7.26\times10^{8}$, $2.43\times10^{36}$, and $2.26\times10^{34}$. The remaining dataset has mean MSE 0.106. The XXZ study likewise contains one selected seed with MSE 31.3. These outcomes remain in the reported aggregates rather than being filtered after test evaluation.

\subsection{History and observation interventions}
\label{app:input-effects}

\begin{table}[h]
\caption{Dataset-level contrasts in Figure~\ref{fig:reversals}. Each dataset column averages three training seeds; the final column averages datasets. Positive architecture differences favor causal attention, and positive input differences favor full history.}
\label{tab:reversal-effects}
\centering
\begin{tabular}{llrrrr}
\toprule
Contrast & Condition & Dataset 1 & Dataset 2 & Dataset 3 & Mean \\
\midrule
LSTM minus causal & 64 trajectories & +0.01292 & +0.00508 & +0.00363 & +0.00721 \\
LSTM minus causal & 256 trajectories & -0.02696 & -0.01586 & -0.02314 & -0.02198 \\
Local minus full input & Observed-history & +0.01967 & +0.02011 & +0.02162 & +0.02046 \\
Local minus full input & Closed-loop & -0.03221 & -0.03072 & -0.02776 & -0.03023 \\
\bottomrule
\end{tabular}
\end{table}

\begin{table}[h]
\caption{Direction summaries across the 32 correlated evaluation views. ``All-data agreement'' counts views in which the sign is the same for all three independently generated datasets.}
\label{tab:history-summary}
\centering
\begin{tabular}{llrr}
\toprule
System & Comparison & Lower-error views & All-data agreement \\
\midrule
Two-spin Ising & LSTM history 5 vs. 20 & 27/32 & 9/32 \\
Four-spin Ising & LSTM history 5 vs. 20 & 30/32 & 28/32 \\
Four-spin Ising & Local vs. full, observed history & 0/16 & 16/16 worse \\
Four-spin Ising & Local vs. full, closed loop & 16/16 & 16/16 better \\
\bottomrule
\end{tabular}
\end{table}

The local-input rows aggregate the direction across LSTM, PatchTST, and causal Transformer models and available histories 1, 5, and 20. The target remains full width. The intervention therefore changes only what state history the predictor receives, not what it must predict or which future controls it knows.

\subsection{Constraint directions}
\label{app:constraint-effects}

\begin{table}[h]
\caption{Two-spin LSTM trace-penalty direction counts on the prespecified constraint dataset.}
\label{tab:constraint-summary}
\centering
\begin{tabular}{rrr}
\toprule
Penalty weight & Views with lower trace error & Views with higher prediction MSE \\
\midrule
0.001 & 31/32 & 26/32 \\
0.01  & 32/32 & 17/32 \\
0.1   & 32/32 & 21/32 \\
\bottomrule
\end{tabular}
\end{table}

For the combined trace, positivity, and bounds objective, prediction MSE increases in all 32 views for both the LSTM and causal Transformer at all three weights. Because constraint calculation is included in the fixed optimization clock, this comparison measures the deployed objective under a fixed resource budget rather than a pure regularization effect at an equal number of gradient updates.

\subsection{Calibration construction}
\label{app:calibration}

\begin{table}[h]
\caption{Nominal 90\% whole-trajectory-region interval results for the two-spin Ising, 256-trajectory, full-history condition. Width is the full symmetric interval width.}
\label{tab:calibration-exact}
\centering
\begin{tabular}{lrrr}
\toprule
Predictor & ID coverage & Shift coverage & Width \\
\midrule
LSTM & 87.50\% & 28.13\% & 1.2767 \\
PatchTST & 91.32\% & 26.91\% & 7.6545 \\
Causal Transformer & 86.98\% & 16.49\% & 1.7195 \\
Low-rank linear & 87.67\% & 24.31\% & 2.3278 \\
\bottomrule
\end{tabular}
\end{table}

For each fixed predictor and region, calibration trajectory \(j\) receives score
\begin{equation}
s_j=\max_{(t,k)\in\mathcal{I}}\left|x_{t,k}^{(j)}-\widehat{x}_{t,k}^{(j)}\right|,
\end{equation}
where \(\mathcal{I}\) is the Cartesian product of reported times and nonidentity channels. With 64 calibration trajectories and target miscoverage \(\alpha=0.1\), the radius is the \(\lceil(64+1)(1-\alpha)\rceil=59\)-th ordered score. The interval uses the same radius for every coordinate in the region and is not clipped. ID and shifted tests share this radius; only the test distribution changes.

\section{Failure-Preserving Diagnostics}
\label{app:failures}

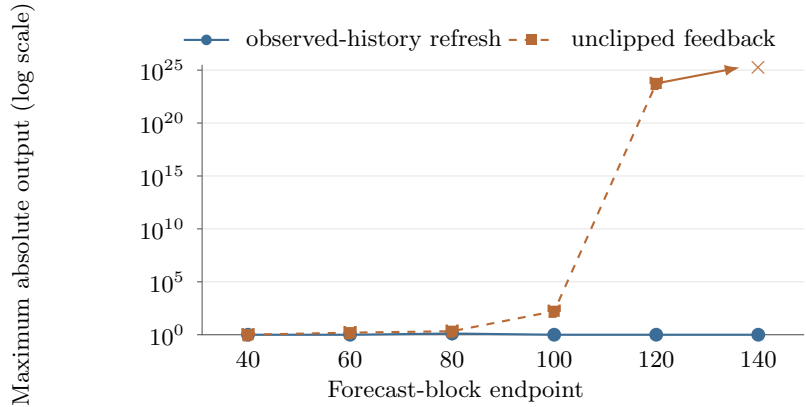
\begin{figure}[!b]
\centering
\begin{tikzpicture}[font=\footnotesize]
\path[use as bounding box] (-0.1,0) rectangle (11.3,5.25);
\node[font=\small,rotate=90] at (0.12,2.65) {Maximum absolute output (log scale)};
\node[font=\small] at (5.85,0.18) {Forecast-block endpoint};
\begin{scope}[shift={(1.75cm,0.92cm)},x=1.35cm,y=0.14cm]
\foreach \y/\tick in {0/{10^{0}},5/{10^{5}},10/{10^{10}},15/{10^{15}},20/{10^{20}},25/{10^{25}}} {
  \draw[black!10] (0.55,\y) -- (6.45,\y);
  \draw[black!65] (0.55,\y) -- ++(-0.05,0);
  \node[anchor=east,xshift=-3pt] at (0.55,\y) {\(\tick\)};
}
\draw[black!65] (0.55,25.5) -- (0.55,0) -- (6.45,0);
\foreach \x/\tick in {1/40,2/60,3/80,4/100,5/120,6/140} {
  \draw[black!65] (\x,0) -- ++(0,-0.55);
  \node[anchor=north,yshift=-3pt] at (\x,0) {\tick};
}
\draw[datasetblue,thick] plot[mark=*,mark size=2.2pt] coordinates {(1,0.0000) (2,0.0000) (3,0.1128) (4,0.0000) (5,0.0000) (6,0.0000)};
\draw[datasetorange,thick,dashed] plot[mark=square*,mark size=2.1pt] coordinates {(1,0.0000) (2,0.1954) (3,0.3305) (4,2.2183) (5,23.7465)};
\draw[datasetorange,thick,-{Latex[length=2mm]}] (5,23.7465) -- (5.82,25.25);
\node[text=datasetorange,font=\bfseries] at (6,25.25) {\(\times\)};
\end{scope}
\draw[datasetblue,thick] (2.25,4.82) -- (2.85,4.82);
\fill[datasetblue] (2.55,4.82) circle (2pt);
\node[anchor=west] at (2.95,4.82) {observed-history refresh};
\draw[datasetorange,thick,dashed] (6.55,4.82) -- (7.15,4.82);
\node[fill=datasetorange,minimum size=4pt,inner sep=0pt] at (6.85,4.82) {};
\node[anchor=west] at (7.25,4.82) {unclipped feedback};
\end{tikzpicture}
\caption{Closed-loop amplification for one prespecified PatchTST model--trajectory pair. Each point is the maximum absolute model output in a successive 20-step block; the vertical axis is logarithmic. Observed-history refresh stays bounded, whereas unclipped feedback reaches \(5.58\times10^{23}\) before the next block becomes nonfinite (cross). This numerical diagnostic is not a population failure-rate estimate.}
\label{fig:closed-loop-failure}
\end{figure}

For one prespecified PatchTST model--trajectory pair, unclipped feedback becomes nonfinite after five completed forecast blocks. Figure~\ref{fig:closed-loop-failure} shows the preceding amplification: the maximum absolute output grows from 2.14 to 165 and then to \(5.58\times10^{23}\), while observed-history refresh remains near one. Replaying the offending input in double precision moves the first observed nonfinite value from the input projection to the output projection but does not restore a finite forecast. Output clipping permits completion, with MSE 0.21652 compared with 0.01082 under observed-history refresh. These are distinct protocol outcomes for a single diagnostic case, not an estimated failure rate.

The current-state MLP analysis includes the finite four-spin and XXZ high-error outcomes listed in Appendix~\ref{app:baselines}. Validation error alone selects their checkpoints and learning rates; stable test behavior is not an eligibility condition.

A paired six-fit diagnostic compares BatchNorm with LayerNorm under shared parameter initialization and matched GPUs on one dataset. BatchNorm-minus-LayerNorm validation errors have mixed signs across seeds and budgets, so the diagnostic does not support a normalization-wide advantage or explain the closed-loop failure. The primary PatchTST comparisons use effective LayerNorm.

One validation-selected predictor cannot initialize on the evaluation device, so its eight protocol--region intervals remain unavailable. No successful training seed is substituted, and summaries requiring complete coverage report the missing value.

\section{Implementation and Reproducibility}

Source code, model and training configurations, dataset-generation metadata, checkpoints, and per-evaluation metrics are organized by experimental condition. Each result identifies its model configuration, dataset seed, training seed, optimization budget, validation-selected checkpoint, metric values, and failure status. This structure supports reconstruction of the aggregation hierarchy and the distinction between missing, nonfinite, clipped, and successfully completed outcomes.

The experiment software is implemented in Python with PyTorch, NumPy, and SciPy. Reference quantum propagation uses SciPy matrix exponentiation for two- and four-spin systems and sparse \texttt{expm\_multiply} for six-spin systems. Machine-readable environment files record the package versions used for data generation, training, and evaluation.

\end{document}